# Combining multiple matchers for fingerprint verification: A case study in Biosecure Network of Excellence

Fernando ALONSO-FERNANDEZ*, Julian FIERREZ-AGUILAR*,
Hartwig FRONTHALER**, Klaus KOLLREIDER**, Javier ORTEGA-GARCIA*,
Joaquin GONZALEZ-RODRIGUEZ*, Josef BIGUN**

**Abstract**

*We report on experiments for the fingerprint modality conducted during the First BioSecure Residential Workshop. Two reference systems for fingerprint verification have been tested together with two additional non-reference systems. These systems follow different approaches of fingerprint processing and are discussed in detail. Fusion experiments involving different combinations of the available systems are presented. The experimental results show that the best recognition strategy involves both minutiae-based and correlation-based measurements. Regarding the fusion experiments, the best relative improvement is obtained when fusing systems that are based on heterogeneous strategies for feature extraction and/or matching. The best combinations of two/three/four systems always include the best individual systems whereas the best verification performance is obtained when combining all the available systems.*

**Key words:** Biometrics, Pattern recognition, Fingerprint, Comparative study, Case study, Identification, Research program, Experimental study, Performance evaluation, Mixed method, Data fusion.

## COMBINAISON DE PLUSIEURS CLASSIFIEURS POUR LA VÉRIFICATION D'EMPREINTES DIGITALES : UNE ÉTUDE DE CAS DANS LE CADRE DU RÉSEAU D'EXCELLENCE BIOSECURE

**Résumé**

*Voici le rapport sur les expériences menées sur la modalité d'empreintes digitales pendant le premier atelier BioSecure Residential. Deux systèmes de référence pour la vérification d'empreinte digitale ont été examinés ainsi que deux systèmes additionnels sans*

* ATVS, Escuela Politecnica Superior, Universidad Autonoma de Madrid – Avda. Francisco Tomas y Valiente 11, Campus de Cantoblanco, 28049 Madrid, Spain;
{fernando.alonso, julian.fierrez, javier.ortega, joaquin.gonzalez}@uam.es
**Halmstad University, SE-30118, Sweden; {hartwig.fronthaler, klaus.kollreider, josef.bigun}@ide.hh.se

*référence. Ces systèmes suivent différentes approches pour le traitement des empreintes digitales, que nous allons présenter en de plus amples détails. En outre, nous présentons des expériences de fusion comportant différentes combinaisons de systèmes disponibles. Les résultats expérimentaux prouvent que la meilleure stratégie d'identification implique des mesures basées sur les minuties et la corrélation. Concernant les expériences de fusion, la meilleure amélioration relative est obtenue en fusionnant les systèmes basés sur des stratégies hétérogènes dans leur extraction de paramètres et/ou dans leur technique de comparaison. Les meilleures combinaisons de deux, trois, ou quatre systèmes incluent toujours le meilleur système. Le meilleur résultat en vérification est obtenu en combinant tous les systèmes disponibles.*

**Mots clés :** Biométrie, Reconnaissance forme, Empreinte digitale, Étude comparative, Étude de cas, Identification, Programme recherche, Étude expérimentale, Évaluation de performance, Méthode mixte, Fusion d'informations.

## Contents

# I. INTRODUCTION

There is an increasing need for reliable automatic personal identification due to the expansion of the networked society. This has resulted in the popularity of *biometrics* [1], which refers to the automatic recognition of individuals based on their physiological and/or behavioral characteristics such as their fingerprint, face, iris, voice, hand, signature, and so on. A wide variety of applications require reliable personal recognition schemes to either confirm or to determine the identity of an individual requesting their services. However, biometric-based systems have some limitations that may have adverse implications for the security of a system and also there are a number of privacy concerns raised about the use of biometrics [1, 2]. The widespread adoption of biometric technologies has proved to be slower than predicted, especially in Europe.

With the purpose of overcoming the current impediments and limitations in existing biometrics systems and increasing the trust and confidence in biometric solutions, the BioSecure Network of Excellence (NoE) [3] was launched in June 2004 within the Information Society Technology priority of the 6[th] Framework Programme of the European Union. During the First Biosecure Residential Workshop [4], different reference systems and research prototypes of several modalities were tested on common databases. In this paper, we report the experiments carried out for the fingerprint modality. Two reference systems for the fingerprint modality are tested and compared with two additional non-reference systems. The four systems tested include different approaches for feature extraction, fingerprint alignment and fingerprint matching. Fusion experiments using standard fusion approaches are also reported.

This paper is structured as follows. Section II describes the reference systems for the fingerprint modality. Section III describes the two additional non-reference systems used in the experiments. Experimental setup and results are given in Section IV. Conclusions are finally drawn in Section V.

## II. BIOSECURE REFERENCE SYSTEMS FOR FINGERPRINT MODALITY

The reference systems for fingerprint modality in the First Biosecure Residential Workshop were: (i) the fingerprint recognition software developed by Halmstad University (HH) in Sweden [5], and (ii) the open-source NIST Verification Test Bed (VTB) [6]. These two reference systems are briefly described below.

### II.1. Minutiae-based fingerprint verification system using complex filtering

The fingerprint recognition software developed by Halmstad University [5] includes a novel way to detect the minutia points' position and direction, and also ridge orientation, by using filters sensitive to parabolic and linear symmetries. The minutiae are exclusively used for alignment of two fingerprints. The number of paired minutiae can be low, which is advantageous in partial or low quality fingerprints. After a global alignment, a matching is performed by distinctive area correlation, involving the minutiae's neighborhood. We briefly describe the four phases of the system, (i) local feature extraction, (ii) pairing of minutiae, (iii) fingerprint alignment, and (iv) matching.

#### II.1.1. Local feature extraction

Two prominent minutia types, ridge bifurcation and termination have parabolic symmetry properties [7], whereas they lack linear symmetry [8]. The leftmost image in Figure 1 shows a perfectly parabolic pattern. On the contrary, the local ridge and valley structure is linearly symmetric. A perfectly linear pattern is a planar wave having the same orientation at each point in a neighborhood.

Averaging the orientation tensor $z = (f_x + if_y)^2$ of an image (with $f_x$ and $f_y$ as its partial derivatives) gives an orientation estimation and its error. Linear symmetry $LS$ is computed by dividing averaged $z$ with averaged $|z|$. The result is a complex number, having the ridge orientation (in double angle) as argument and the reliability of its estimation as magnitude. Parabolic symmetry $PS$ is retrieved by convolving $z$ with a filter $h_n = (x + iy)^n \cdot g$, where g denotes a 2D Gaussian, with $n = 1$. The result is again a complex number, having the minutiae's direction as argument and an occurrence certainty as magnitude (compare Figure 1). Note, that $h_0$ can be used for the calculation of $LS$. All filtering is done in 1D involving separable Gaussians and their derivatives.

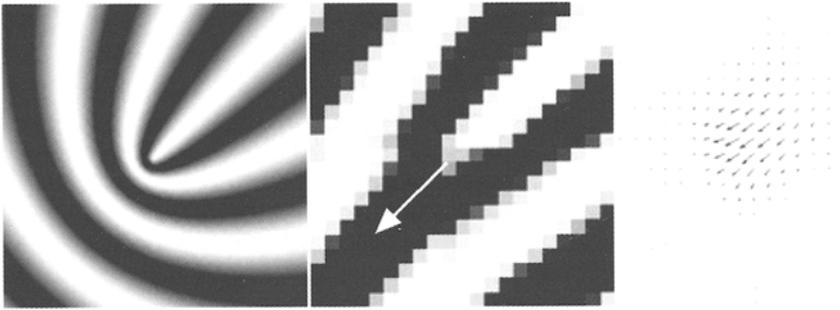

FIG 1. – Left-hand side: perfectly parabolic pattern; center: ridge bifurcation neighborhood with indicated minutia direction; right-hand side: corresponding complex response of $h_1$ when convoluted with z.

*Côté gauche: modèle parfaitement parabolique; centre: voisinage de bifurcation d'arête avec la direction indiquée de minutie; côté droit: réponse complexe correspondante de $h_1$ en faisant la convolution avec z.*

At the beginning, an image enhancement [9] is applied prior to the calculation of linear and parabolic symmetries. Then, some additional measures are taken in order to reliably detect minutiae points. First, the selectivity of the parabolic symmetry filter response is improved, using a simple inhibition scheme to get $PSi = PS \cdot (1 - |LS|)$. Basically, parabolic symmetry is attenuated if the linear symmetry is high, whereas it is preserved in the opposite case. In Figure 2, the overall minutia detection process is depicted. The first two images show the initial fingerprint and its enhanced version, respectively. The extracted parabolic symmetry is displayed in image IV ($|PS|$), whereas the linear part is shown in image III ($LS$). The sharpened magnitudes $|PSi|$ are displayed in image V.

To avoid multiple detections of the same minutia, neighborhoods of $9 \times 9$ pixels are considered when looking for the highest responses in $PSi$. At this stage, $LS$ can be reused to verify minutia candidates: Firstly, a minimum $|LS|$ is employed to segment the fingerprint area from the image background. Secondly, each minutia is required to have full surround of high linear symmetry, in order to exclude spurious and false minutiae. Minutiae's coordinates and direction are stored in a list ordered by magnitude. In image VI of Figure 2, its first 30 entries are indicated by circles.

### II.1.2. Pairing of minutiae

In order to establish correspondences between two fingerprints, a local minutia matching approach inspired by triangular matching [10] is implemented. This essentially means establishing a connected series of triangles, which are equal with respect to both fingerprints and have corresponding minutiae as their corners.

For each minutia in a fingerprint, additional attributes are derived, which describe their within-fingerprint relation. For two arbitrary minutiae $m_i$ and $m_j$ of one fingerprint, the following attributes are derived: (i) the distance $d_{ij} = d_{ji}$ between the two minutiae; and (ii) the angles $\alpha_{ij}$ and $\alpha_{ji}$ of the minutiae with respect to the line between each other (compare Figure 3). Next, corresponding couples in the two fingerprints are selected. Having two arbi-

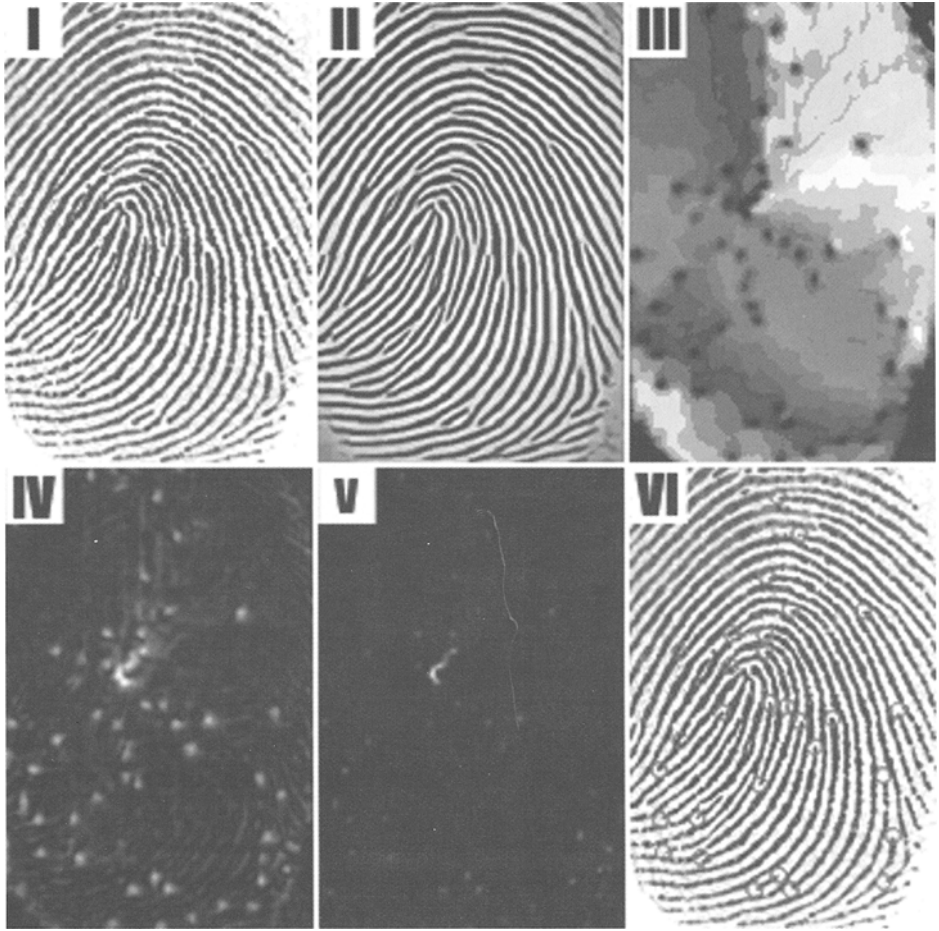

FIG 2. – Local feature extraction using complex filtering (HH reference system);
image III: linear symmetry LS; image IV: parabolic symmetry PS.

*Extraction locale de paramètres utilisant un filtrage complexe (système de référence de HH);
image III: symétrie linéaire LS; image IV: symétrie parabolique PS.*

trary minutiae $m_k$ and $m_l$ of the second fingerprint, the correspondence is fulfilled if $|d_{ij} - d_{kl}| < \lambda_{dist}$ and $(|\alpha_{ij} - \alpha_{kl}| + |\alpha_{ji} - \alpha_{lk}|) < \lambda_{angle}$. Thus, a corresponding couple means two pairs of minutiae, e.g. $\{m_i, m_j; m_k, m_l\}$, which at least correspond in a local scope.

Among all corresponding couples, we look for those which have a minutia in common in both of the fingerprints. Taking $\{m_i, m_j; m_k, m_l\}$ as a reference, it may be that $\{m_i, m_o; m_k, m_p\}$ and $\{m_j, m_o; m_l, m_p\}$ are corresponding couples as well. This is also visualized right, in Figure 3. Such a constellation suggests $m_o$ and $m_p$ being neighbors to $\{m_i, m_j\}$ and $\{m_k, m_l\}$, respectively. To verify neighbors, we additionally check the closing angles $\gamma_1$ and $\gamma_2$ in order to favor uniqueness. In this way neighbors are consecutively assigned to the corresponding

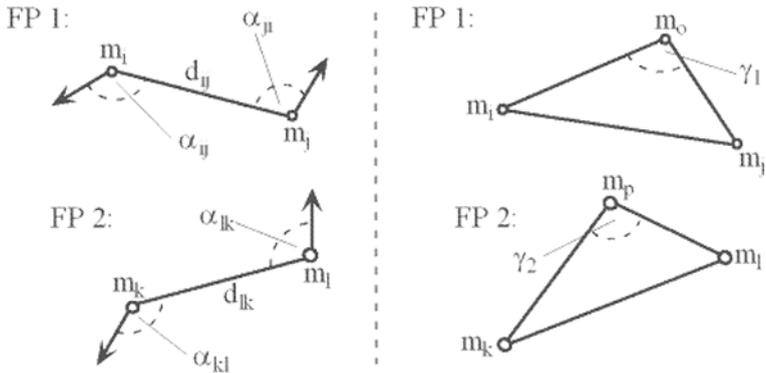

FIG 3. – Corresponding couples (left-hand side) and triangles (right-hand side) for two fingerprints; all angles are signed in order to be unambiguous.

*Couples (côté gauche) et triangles (côté droit) correspondants pour deux empreintes digitales ; tous les angles sont indicés afin d'éviter toute ambiguïté.*

reference couples, the equivalent of establishing equal triangles with respect to both fingerprints sharing a common side. Each corresponding couple is taken as a reference once. The corresponding couple to which most neighbors can be found is considered for further processing. This couple as well as its mated neighbors is stored in a pairing list.

### II.1.3. Fingerprint alignment

Here, global alignment of two fingerprints is assumed to be a rigid transformation since only translation and rotation is considered. The corresponding parameters are computed using the established minutia pairs (list): Translation is given by the difference of the position vectors for the first minutia pair. The rotation parameter is determined as the averaged angle among vectors between the first minutia pair and all others. Following the estimated parameters, the coordinate transformation for all points in $LS$ is done, as the latter is needed for the final step. No further alignment efforts, e.g. fine adjustment, are done.

### II.1.4. Fingerprint matching

Finally, a simple matching using normalized correlation at several sites of the fingerprint is done (similar to [11]). Small areas in $LS$ around the detected minutia points in the first fingerprint are correlated with areas at the same position in the second fingerprint. Only areas having an average linear symmetry higher than a threshold are considered. This is done to favor well-defined (reliable) fingerprint regions for comparison. The final matching score is given by the mean value of the single similarity measures.

## II.2. NIST Fingerprint Image Software (NFIS2)

The NIST Fingerprint Image Software 2 (NFIS2) contains software technology, developed for the Federal Bureau of Investigation (FBI), designed to facilitate and support the automated manipulation and processing of fingerprint images. Source code for over 50 different utilities or packages and an extensive User's Guide are distributed on CD-ROM free of charge [12].

For our evaluation and tests with NFIS2, we have used the following packages: (i) MINDTCT for minutiae extraction; and (ii) BOZORTH3 for fingerprint matching. These two packages are described next.

### II.2.1. Minutiae extraction using MINDTCT

MINDTCT takes a fingerprint image and locates all minutiae in the image, assigning to each minutia point its location, orientation, type, and quality. The algorithms used in MINDTCT were inspired by the Home Office's Automatic Fingerprint Recognition System; specifically the suite of algorithms commonly referred to as "HO39" [13]. The architecture of MINDTCT is shown in Figure 4 and it can be divided in the following phases: (i) generation of image quality map; (ii) binarization; (iii) minutiae detection; (iv) removal of false minutiae, including islands, lakes, holes, minutiae in regions of poor image quality, side minutiae, hooks, overlaps, minutiae that are too wide, and minutiae that are too narrow (pores); (v) counting of ridges between a minutia point and its nearest neighbors; and (vi) minutiae quality assessment. Additional details of these phases are given below.

Because of the variation of image quality within a fingerprint, NFIS2 analyzes the image and determines areas that are degraded. Several characteristics are measured, including regions of low contrast, incoherent ridge flow, and high curvature. These three conditions represent unstable areas in the image where minutiae detection is unreliable, and together they are used to represent levels of quality in the image. An image quality map is generated integrating these three characteristics. Images are divided into non-overlapping blocks, where one out of five levels of quality is assigned to each block.

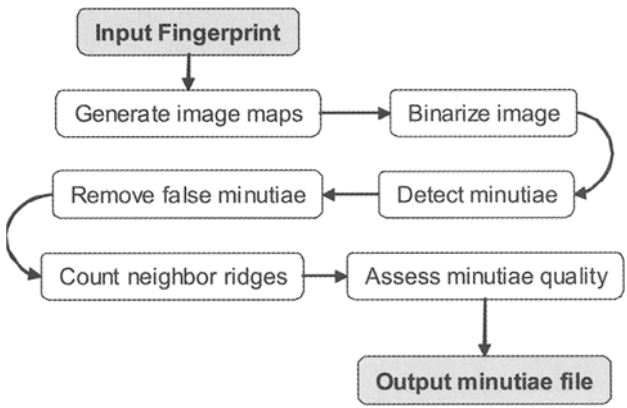

FIG 4. – System architecture of the MINDTCT package of the NIST Fingerprint Image Software 2 (NFIS2)

*Architecture de système du paquetage MINDTCT du logiciel NIST Fingerprint Image Software 2 (NFIS2)*

The minutiae detection step scans the binary image of the fingerprint, identifying local pixel patterns that indicate the ending or splitting of a ridge. A set of minutia patterns is used to detect candidate minutia points. Subsequently, false minutiae are removed and the remaining candidates are considered as the true minutiae of the image. Fingerprint minutiae matchers often use information in addition to just the points themselves. Apart from minutia's position, direction, and type, MINDTCT computes ridge counts between a minutia point and each of its nearest neighbors.

In the last step, a quality/reliability measure is associated with each detected minutia point. Even after performing the removal step, false minutiae potentially remain in the list. A robust quality measure can help to manage this. Two factors are combined to produce a quality measure for each detected minutia point. The first factor is taken directly from the location of the minutia point within the quality map described before. The second factor is based on simple pixel intensity statistics (mean and standard deviation) within the immediate neighborhood of the minutia point. A high quality region within a fingerprint image is expected to have significant contrast that will cover the full grayscale spectrum [12].

### II.2.2. Fingerprint matching using BOZORTH3

The BOZORTH3 matching algorithm computes a match score between the minutiae from any two fingerprints to help determine if they are from the same finger. The BOZORTH3 matcher uses only the location and orientation of the minutia points to match the fingerprints. It is rotation and translation invariant. The algorithm can be described by the following three steps: (i) construction of two Intra-Fingerprint Minutia Comparison Tables, one table for each of the two fingerprints; (ii) construction of an Inter-Fingerprint Compatibility Table; and (iii) generation of the matching score using the Inter-Fingerprint Compatibility Table. These steps are described below.

The first step is to compute relative measurements from each minutia in a fingerprint to all other minutia in the same fingerprint. These relative measurements are stored in the Intra-Fingerprint Minutia Comparison Table and are used to provide rotation and translation invariance. The invariant measurements computed are distance between two minutiae and angle between each minutia's orientation and the intervening line between both minutiae. A comparison table is constructed for each of the two fingerprints.

The next step is to take the Intra-Fingerprint Minutia Comparison Tables from the two fingerprints and look for "compatible" entries between the two tables. Table entries are "compatible" if: (i) the corresponding distances and (ii) the relative minutia angles are within a specified tolerance. An Inter-Fingerprint Compatibility Table is generated, only including entries that are compatible. A compatibility table entry therefore incorporates two pairs of minutia, one pair from the template fingerprint and one pair from the test fingerprint. The entry into the compatibility table indicates that the minutiae pair of the template fingerprint corresponds to the minutiae pair of the test fingerprint.

At the end of the second step, we have constructed a compatibility table which consists of a list of compatibility association between two pairs of potentially corresponding minutiae. These associations represent single links in a *compatibility graph*. The matching algorithm then traverses and links table entries into clusters, combining compatible clusters and accumulating a match score. The larger the number of linked compatibility associations, the higher the match score, and the more likely the two fingerprints originate from the same person.

## III. OTHER NON-REFERENCE SYSTEMS

We have also tested the minutiae-based [14] and the ridge-based [15] fingerprint matchers developed in the Biometrics Research Lab. at Universidad Politecnica de Madrid (UPM), Spain.

### III.1. UPM minutiae-based fingerprint verification system

The UPM minutiae matcher is described in [14]. Its architecture is shown in Figure 5 and it can be divided in three phases: (i) image enhancement, in order to reconstruct the ridge structure of the fingerprint; (ii) feature extraction; and (iii) pattern matching process, where a fingerprint sample is compared with a pattern registered in the database.

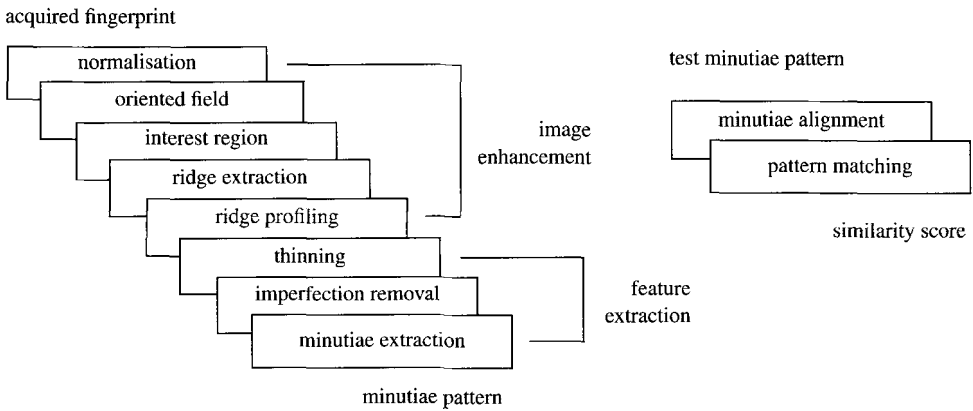

FIG 5. – System architecture of the UPM-minutiae based verification system.

*L'architecture du système de vérification basé sur les minuties UPM.*

#### III.1.1. Image enhancement

This stage is aiming to provide a high-quality image, so that the feature extractor can obtain a precise biometric pattern. In our system, we use the minutiae pattern. A point in the fingerprint image is considered as minutia if it is derived from an ending, beginning or bifurcation of a ridge. Image imperfections may induce errors in the estimation of the spatial coordinates and relative orientation of each minutia, thus decreasing the reliability of the recognition system. This makes the image enhancement step necessary. The complete sequence of stages performed in our system is: (i) normalization; (ii) computation of the orientation field; (iii) region of interest extraction; (iv) ridge extraction; and (v) ridge profiling. Details of each stage are explained in [16, 17].

### III.1.2. Feature extraction

This stage is aiming to provide a reliable biometric pattern of the fingerprint. The sequence of stages performed in our system is [16, 17, 18]: (i) thinning of the enhanced binary ridge structure resulting from the image enhancement; (ii) removal of imperfections from the thinned image; and (iii) minutiae extraction. For each detected minutia the following parameters are stored: (i) the x and y coordinates of the minutia; (ii) the orientation angle of the ridge containing the minutia; (iii) in the case of an ending ridge, the x and y co-ordinates of the sampled ridge segment containing the minutia; (iv) in the case of a ridge bifurcation, the x and y co-ordinates of the sampled ridge segment of one of the bifurcation branches. A sample fingerprint image, the resulting binary fingerprint (after image enhancement), the minutiae pattern superimposed on the thinned image and its corresponding sampled ridge segments are shown in Figure 6.

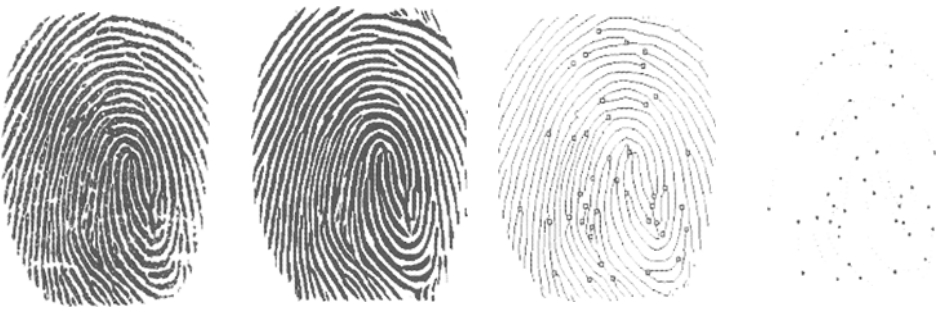

FIG 6. – Processing steps of the UPM-minutiae based verification system.

*Les étapes de transformation du système de vérification basé sur les minuties UPM.*

### III.1.3. Fingerprint matching

The aim of this stage is to determine whether two biometric patterns have been produced by the same finger or not. Due to elastic deformations introduced in the acquisition process and imperfections in the image, the two patterns must be aligned before matching. Pattern alignment is achieved considering the relative position of the minutiae in the image, and then the matching process is accomplished seeking correspondence between the two aligned structures. In our matching process, a score is defined to measure the similarity (edit distance) between the compared patterns. An elastic technique for minutiae comparison is used, permitting certain spatial tolerance margin. To compensate for the nonlinear elastic deformation of the skin, the technique is also adaptive. For this purpose, a size-adaptive tolerance box adjustable to the spatial co-ordinates values of the explored minutiae is defined.

## III.2. UPM ridge-based fingerprint verification system

The UPM ridge-based matcher uses a set of Gabor filters to capture the ridge strength. The image is tessellated into square cells, and the variance of the filter responses in each cell across all filtered images is used as feature vector. This feature vector is called FingerCode because of the similarity to previous research works [19]. The automatic alignment is based on the system described in [20], in which the correlation between the two FingerCodes is computed, obtaining the optimal offset. The UPM ridge-based matcher can be divided in two phases: (i) extraction of the FingerCode; and (ii) matching of the FingerCodes.

### III.2.1. Extraction of the FingerCode

This stage is aiming to provide a reliable biometric pattern of the fingerprint. No image enhancement is performed since Gabor filters are robust enough to the typical noise present in the fingerprint images. The complete sequence of stages performed in our system is: (i) convolution of the input fingerprint image with 8 Gabor filters, obtaining 8 filtered images $F_\theta$; (ii) tessellation of the filtered images into equal-sized square disjoint cells; and (iii) extraction of the FingerCode. For each cell of each filtered image $F_\theta$, we compute the variance of the pixel intensities. These standard deviation values constitute the FingerCode of a fingerprint image. A sample fingerprint image, the resulting convolved image with a Gabor filter of orientation $\theta = 0°$, the tessellated image and its FingerCode are shown in Figure 7.

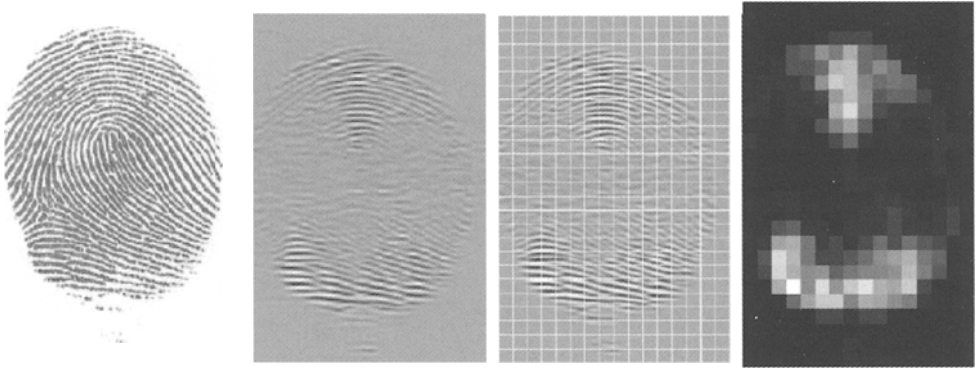

FIG 7. – Processing steps of the UPM-ridge based verification system.

*Les étapes de transformation du système de vérification basé sur les sillons UPM.*

### III.2.2. Matching of the FingerCodes

The aim of this stage is to determine whether two biometric patterns have been produced by the same finger or not. The complete sequence of stages performed in our system is: (i) alignment of the two fingerprints to be compared; (ii) re-computation of the FingerCodes of the two fingerprints once they have been aligned; and (iii) matching between the re-computed FingerCodes. The matching score is computed as the Euclidean distance between the two FingerCodes.

To determine the alignment between two fingerprints, we compute the 2D correlation of the two FingerCodes [20]. Correlation involves multiplying corresponding entries between the two FingerCodes at all possible translation offsets, and determining the sum, which is computed more efficiently in the Fourier domain. The offset that results in the maximum sum is chosen to be the optimal alignment. Every offset is properly weighted to account for the amount of overlap between the two FingerCodes. Worth noting, this procedure does not account for rotational offset between the two fingerprints. For the standard database MCYT used in this work, which is acquired under realistic conditions with an optical sensor, we have observed that typical rotations between different impressions of the same fingerprint are compensated for by using tessellation.

## IV. EXPERIMENTS

### IV.1. Database and experimental protocol

A large biometric database acquisition process was launched in 2001 within the MCYT project [21]. A single-session fingerprint database acquisition was designed to include different types of sensors and different acquisition conditions. Two types of acquisition devices were used: (i) a CMOS-based capacitive capture device, model 100SC from Precise Biometrics producing a 500 dpi, 300 × 300 pixel image; and (ii) an optical scanning device, model UareU from Digital Persona, producing a 500 dpi, 256 × 400 pixel image.

With the aim of including variability in fingerprint positioning on the sensor, the MCYT database includes 12 different samples of each fingerprint, all of them acquired under human supervision and considering three different levels of control. For this purpose, the fingerprint core must be located inside a size-varying rectangle displayed in the acquisition software interface viewer. In Figure 8, three samples of the same fingerprint are shown, so that variability in fingerprint positioning can be clearly observed. Depending on the size of the rectangle, the different levels of control will be referred to as: (i) "low", with 3 fingerprint acquisitions using the biggest rectangle; (ii) "medium", with 3 fingerprint acquisitions; and (iii) "high", with 6 fingerprint acquisitions using the smallest rectangle.

Some example images of the MCYT database acquired with the optical sensor are shown in Figure 9.

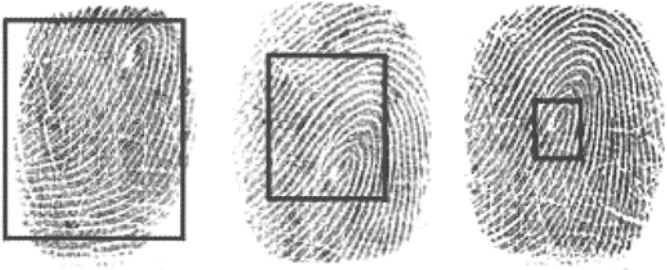

FIG 8. − Examples of the same MCYT fingerprint samples acquired at different levels of control.

*Exemples des mêmes échantillons d'empreintes digitales MYCT acquis à différents niveaux de contrôle.*

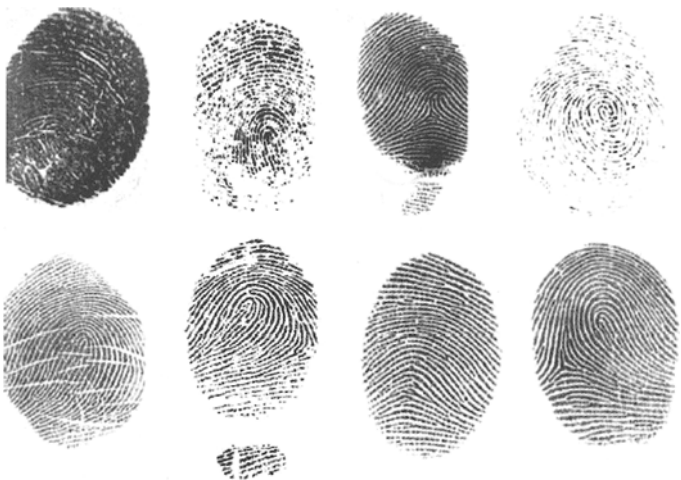

FIG 9. − Several examples of MCYT fingerprints.

*Plusieurs exemples d'empreintes digitales MCYT.*

For the experiments reported in this paper, we use a subcorpus of the MCYT Database. Data consist of 7500 fingerprint images from the 10 fingers of 75 contributors acquired with the optical sensor. We consider the different fingers as different users enrolled in the system, thus resulting in 750 users with 10 impressions per user.

We use one impression per finger with low control during the acquisition as a template. In genuine trials, the template is compared to the other 9 impressions available (2 with low control, 2 with medium control and 5 with high control). The impostor trials are obtained by comparing the template to one impression with high control of the same finger of all the other contributors. The total number of genuine and impostor trials are therefore 750×9=6,750 and 750×74=55,500, respectively.

## IV.2. Evaluation of the individual systems

In Figure 10 we show the verification results of all individual verification systems described in Sections II and III. Furthermore, a general algorithmic description of these individual systems is given in Table I.

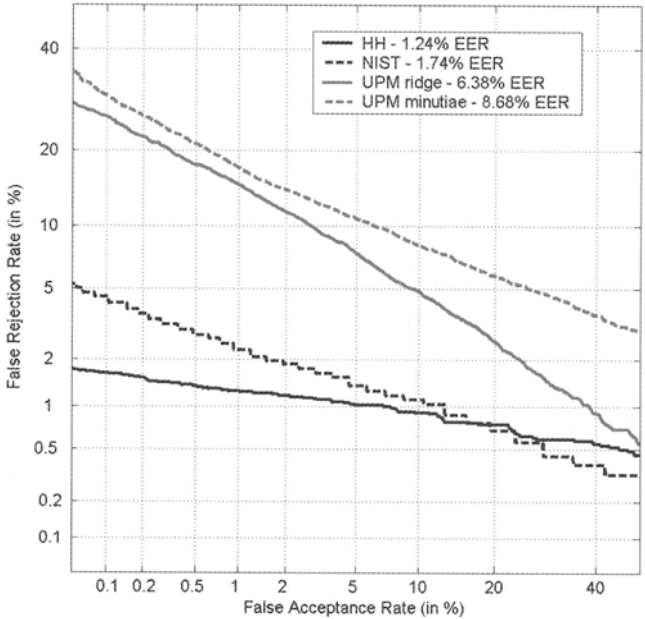

FIG 10. – Verification performance of the individual test systems

*Performance de vérification des systèmes d'essai individuel.*

As expected, we can observe in Figure 10 that minutiae-based matchers perform better than the ridge-based matcher. It is also supported by other findings that minutiae are more discriminative than other features of fingerprints, such as local orientation and frequency, ridge shape or texture information [2]. Only the UPM minutia-based algorithm performs worse than the ridge-based one.

By considering only minutiae-based approaches, HH algorithm results in the best performance. In addition, both HH and NIST algorithms (1.24% and 1.74% EER, respectively) clearly outperform UPM minutiae-based algorithm (8.68% EER). These results may be justified as follows:

TABLE I. – High-level description of the individual systems tested.

*Description à haut niveau des différents systèmes examinés.*

| | Segmentation (Y/N) | Enhancement (Y/N) | Features | Alignment Translation Rotation | | Matching |
|---|---|---|---|---|---|---|
| HH | Y | Y | Minutiae by complex filtering (parabolic and linear symmetry) | TR | | Minutiae-based by triangular matching / Normalized correlation of the neighborhood around minutiae |
| NIST | Y | Y | Minutiae by binarization | TR | | Minutiae-based by comptability between minutiae pairs / Comptability association between pairs of minutiae |
| UPM min | Y | Y | Minutiae by binarization and thining | TR | | Minutiae-based by comptability between individual minutiae / Edit distance between minutiae patterns |
| UPM ridg | N | N | Ridge information by Gabor filtering and square tessellation | T | | Correlation between the extracted features / Euclidean distance between extracted features |

(i) HH algorithm relies on complex filtering for minutiae extraction, considering the surrounding ridge information directly in the gray-scale image. Whereas, NIST and UPM algorithms rely on binarization and morphological analysis, which does not take the surrounding ridge information of the gray-scale image into account, but only the information contained in small neighborhoods of the binarized image. Binarization-based methods usually result in a significant loss of information during the binarization process and in a large number of spurious minutiae introduced during thinning [2], thus decreasing the performance of the system.

(ii) For fingerprint alignment, the UPM algorithm performs matching of individual minutiae, whereas the NIST algorithm matches minutia pairs. The HH algorithm performs triangular matching of minutiae. As the complexity of the alignment method increases, more conditions are implicitly imposed for a fingerprint to be correctly aligned, resulting in higher accuracy.

(iii) In the same way, for fingerprint matching, the UPM algorithm looks for compatibility of individual minutiae, whereas NIST algorithm looks for compatibility of minutiae pairs. Moreover, HH algorithm does not perform minutiae matching but local correlation of areas around the minutiae. Thus, HH algorithm combines the accuracy of a minutiae-based representation with the robustness of a correlation-based matching, which is known to perform properly in low image quality conditions [2].

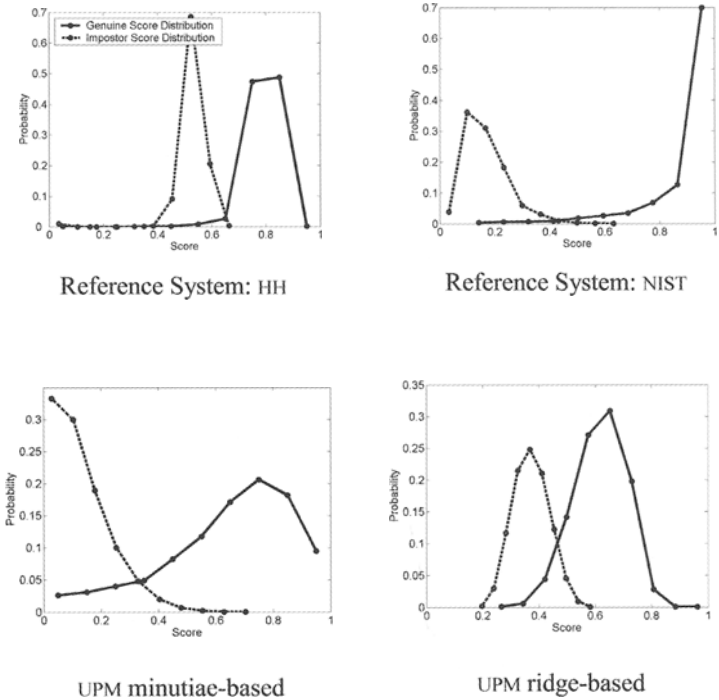

FIG 11. – Genuine and impostor score distributions of the four matchers.
*Score des utilisateurs et des imposteurs des quatres comparateurs.*

## IV.3. Fusion experiments

In this work we have evaluated two different simple fusion approaches based on the max rule and sum rule. These schemes have been used to combine multiple classifiers in biometric authentication with good results reported [22, 23]. The use of these simple fusion rules is motivated by the fact that complex trained fusion approaches do not clearly outperform simple fusion approaches, e.g. see [24].

Figure 11 depicts the genuine and impostor score distributions of the four matchers described in Sections II and III when following the experimental protocol defined in Section IV.1. Each matching score has been normalized to be a similarity score in the [0,1] range by doing the following: (i) the HH algorithm already provides similarity scores in the desired range since it computes a normalized correlation; (ii) the similarity scores provided by the NIST and UPM minutiae-based algorithms are normalized using $tanh\,(s/c)$ ; and (iii) the dissimilarity scores provided by the UPM ridge-based algorithm are normalized using $exp\,(-s/c)$. In the two last normalization schemes, $s$ denotes the raw score, whereas the parameter $c$ has been chosen heuristically for each matcher to evenly distribute the genuine and impostor score distributions into [0,1].

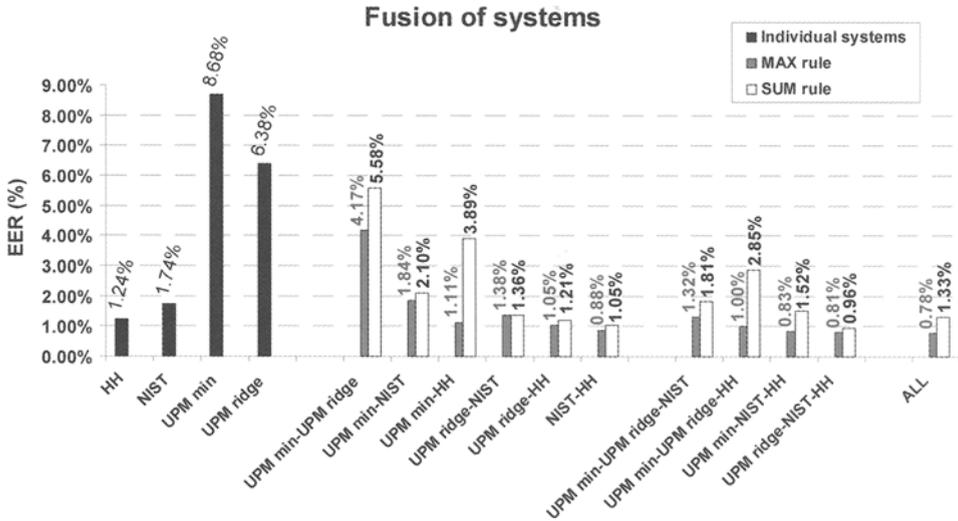

FIG 12. – Verification results of the fusion experiments carried out using the max and sum rule. The first four bars show the results of the individual matchers described in Sections II and III. The remaining bars show the results of all the possible combinations of two, three and four matchers.

*Résultats de vérification des expériences de fusion effectuées en utilisant la règle de maximum et de somme. Les quatre premières barres montrent les résultats des différents comparateurs décrites dans les sections II et III. Les barres restantes montrent les résultats de toutes les combinaisons possibles de deux, trois et quatre comparateurs.*

In Figure 12 we show verification results of the fusion experiments carried out using the max and sum rule. We have evaluated all the possible combinations of two, three and four matchers among the four individual ones described in Sections II and III. Table II indicates in addition the relative performance gain/loss of each combination (max rule only) over the best matcher involved. As can be seen from Figure 12, fusion using the max rule outperforms sum rule based in all cases. Finally, Figure 13 shows verification performance of the best combinations of two, three and four matchers.

From Table II, it is worth noting that the best relative improvement is obtained when fusing systems that are based on heterogeneous strategies for feature extraction and/or matching, thus giving complementary information. For example, HH and NIST algorithms use minutiae as a feature, but they rely on completely different strategies for feature extraction (complex filtering vs. binarization) and matching (normalized correlation vs. minutiae compatibility), see Table I. The same applies for the fusion of UPM minutiae- and UPM ridge-based or NIST and UPM ridge-based algorithms. Interestingly, NIST and UPM minutiae-based algorithms rely on similar strategies for feature extraction and matching, so they are not expected to give complementary information. This may be the reason of the performance worsening observed when fusing these two algorithms.

TABLE II. – Verification results of the fusion experiments carried out using the max and sum rule. The relative performance gain/loss of the max-rule based fusion of algorithms compared to the best matcher involved is also given.

*Résultats de vérification des expériences de fusion effectuées en utilisant la règle de maximum et de somme. La Performance relative gain/pertes de la fusion d'algorithmes basée sur la règle du maximum par rapport au meilleur comparateur est également donnée.*

|  | EER | MAX<br>Relative variation | SUM<br>EER |
|---|---|---|---|
| UPM min-UPM ridge | 4.17% | *(- 34.55%)* | 5.58% |
| UPM min-NIST | 1.84% | (+ 5.74%) | 2.10% |
| UPM min-HH | 1.11% | (- 10.32%) | 3.89% |
| UPM ridge-NIST | 1.38% | (- 20.75%) | 1.36% |
| UPM ridge-HH | 1.05% | (- 15.27%) | 1.21% |
| NIST-HH | 0.88% | *(- 29.22%)* | 1.05% |
| UPM min-UPM ridge-NIST | 1.32% | (-24.22%) | 1.81% |
| UPM min-UPM ridge-HH | 1.00% | (- 18.81%) | 2.85% |
| UPM min-NIST-HH | 0.83% | *(- 32.98%)* | 1.52% |
| UPM ridge-NIST-HH | 0.81% | *(- 34.24%)* | 0.96% |
| ALL | 0.78% | *(- 36.6%)* | 1.33% |

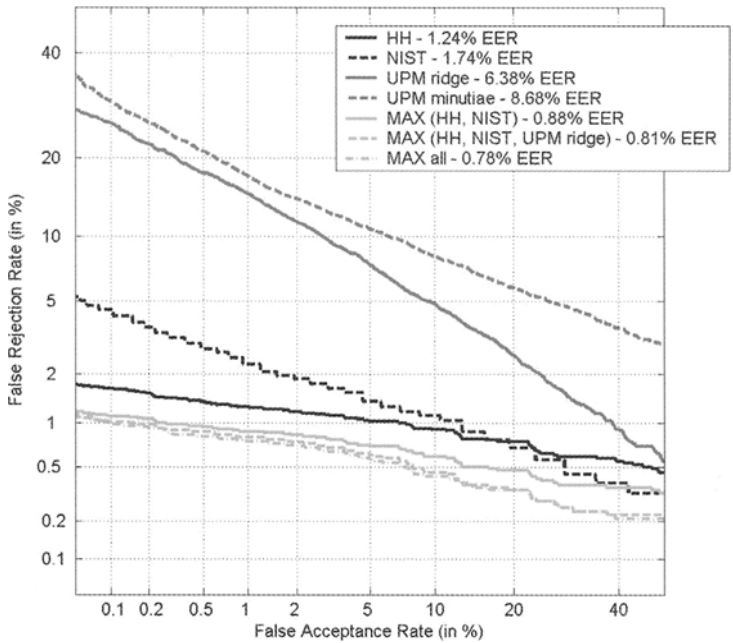

FIG 13. – Verification results for the best combinations of two, three and four matchers. Verification performance of the individual systems is also depicted.

*Résultats de vérification pour les meilleures combinaisons de deux, trois et quatre comparateurs La performance de vérification des différents systèmes est également donnée.*

In terms of EER, the best combination of two systems (HH and NIST) results in a significant performance improvement (1.24% to 0.88% EER). Subsequent inclusion of a third system (UPM ridge-based algorithm) only produces a slightly improvement of the performance (0.88% to 0.81% EER). Interestingly, the best combinations always include the best individual systems (HH and NIST). This should not be taken as a general statement because none of our fusion methods used training. Other studies have revealed that the combination of the best individual systems can be outperformed by other combinations [24] especially if the supervisor is data or expert adaptive. In addition, we also observe that the best EER is obtained when fusing all systems, which also should not be taken as a general statement for the same reasons. In our case, we have a small number of systems and all of them are demonstrated to be in some sense complementary.

## V. DISCUSSION AND CONCLUSIONS

In this work, we have reported on experiments carried out for the fingerprint modality during the First BioSecure Residential Workshop [4]. Two reference systems [5, 6] for fingerprint verification have been tested and compared with two additional non-reference systems [14, 15] on a subcorpus of the MCYT Biometric Database [21]. The four systems implement different approaches for feature extraction, fingerprint alignment and matching. Furthermore, several combinations of the systems using simple fusion schemes have been reported.

A number of experimental findings can be put forward as a result. First, we can confirm that minutiae are more discriminative than other features of the fingerprint, such as local orientation and frequency, ridge shape or texture information [2]. However, methods using alternatives to minutiae-based matching are known to work properly in low image quality conditions [2]. The minutiae-based algorithm that results in the best performance (HH) exploits both a minutiae-based correspondence and a correlation-based matching, instead of using only either of them. Moreover, HH algorithm extracts minutiae by means of complex filtering, instead of using the classical approach based on binarization, which is known to result in loss of information and spurious minutiae [2].

When fusing several systems, the best relative improvement is obtained when combining methods that are based on heterogeneous strategies for feature extraction and/or matching, thus exploiting complementary information. This fact is corroborated in a number of studies [24]. Interestingly, in our experiments, fusing two systems that implement obviously similar strategies leads to a decrease of verification performance. When combining only two systems we generally obtain a significant performance improvement compared to including a third and fourth system. Though the latter combination produces the overall best EER of ~0.8%, it is not the scope of this work to work towards a perfect verification rate but to give an incentive to combine different methods within the same modality and reveal the fundamentals for improvements.

Interestingly, in this study which used untrained supervisors, the best combinations of two/three/four systems always included the best individual systems, and all systems together achieve the best performance. The study [24] reported that combinations of the best individual systems can be outperformed by other combinations if the supervisor is data quality and/or expert adaptive. To check this we would have needed even more systems. Other studies have shown that different individual systems are found to be best depending on the databases' acquisition device [25]. This motivates us to extend the experiments of this work to different databases.

## Acknowledgements

This work has been supported by European FP6 IST-2002-507634 BioSecure NoE and Spanish MCYT TIC2003-08382-C05-01 projects. The authors Fernando Alonso-Fernandez and Julian Fierrez-Aguilar are also supported by a FPI scholarship from Consejeria de Educacion de la Comunidad de Madrid and Fondo Social Europeo.